\documentclass{article}

\PassOptionsToPackage{numbers, compress}{natbib}

 \usepackage[preprint]{neurips_2020}

\usepackage[utf8]{inputenc} 
\usepackage[T1]{fontenc}    
\usepackage[colorlinks,bookmarks=false]{hyperref}
\usepackage{url}            
\usepackage{booktabs}       
\usepackage{amsfonts}       
\usepackage{nicefrac}       
\usepackage{microtype}      
\usepackage{graphicx}
\usepackage{multirow}
\usepackage{amsmath}

\title{Auxiliary Signal-Guided Knowledge Encoder-Decoder for Medical Report Generation}

\author{%
  Mingjie Li \\
  Faculty of Information Technology \\
  Monash University\\
  \texttt{Mingjie.Li@monash.edu}
  \And
  Fuyu Wang \\
  School of Intelligent Systems Engineering \\
  Sun Yat-sen University \\
  \texttt{wangfy8@mail2.sysu.edu.cn} \\
  \AND
  Xiaojun Chang \thanks{Corresponding author.} \\
  Faculty of Information Technology \\
  Monash University \\
  \texttt{cxj273@gmail.com} \\
  \And
  Xiaodan Liang \\
  School of Intelligent Systems Engineering \\
  Sun Yat-sen University \\
  \texttt{xdliang328@gmail.com} \\
}

\begin{document}

\maketitle

\begin{abstract}

Beyond the common difficulties faced in the natural image captioning, medical report generation specifically requires the model to describe a medical image with a fine-grained and semantic-coherence paragraph that should satisfy both medical commonsense and logic. Previous works generally extract the global image features and attempt to generate a paragraph that is similar to referenced reports; however, this approach has two limitations. Firstly, the regions of primary interest to radiologists are usually located in a small area of the global image, meaning that the remainder parts of the image could be considered as irrelevant noise in the training procedure. Secondly, there are many similar sentences used in each medical report to describe the normal regions of the image, which causes serious data bias. This deviation is likely to teach models to generate these inessential sentences on a regular basis. To address these problems, we propose an Auxiliary Signal-Guided Knowledge Encoder-Decoder (ASGK) to mimic radiologists' working patterns. In more detail, ASGK integrates internal visual feature fusion and external medical linguistic information to guide medical knowledge transfer and learning. The core structure of ASGK consists of a medical graph encoder and a natural language decoder, inspired by advanced Generative Pre-Training (GPT). Experiments on the CX-CHR dataset and our COVID-19 CT Report dataset demonstrate that our proposed ASGK is able to generate a robust and accurate report, and moreover outperforms state-of-the-art methods on both medical terminology classification and paragraph generation metrics.

\end{abstract}

\section{Introduction}

Natural image captioning, which aims to summarise visual information (images or videos) in a sentence or generate a topic-related paragraph \cite{anderson2018bottom,rennie2017self,xu2015show}, is a complex task that requires the model to bridge visual and linguistic information. When compared to describing natural images~\cite{cao2018retrieve,wu2016encode}, medical report generation requires an increased capability to understand medical domain knowledge and describe images at a fine-grained and semantic-coherent level, covering accurate abnormal terminologies~\cite{li2018hybrid}. In particular, outstanding challenges associated with modeling medical reports lie in successfully detecting visual groundings and incorporating medical domain knowledge.

\begin{figure}[t]
\centering
\includegraphics[width=1\textwidth]{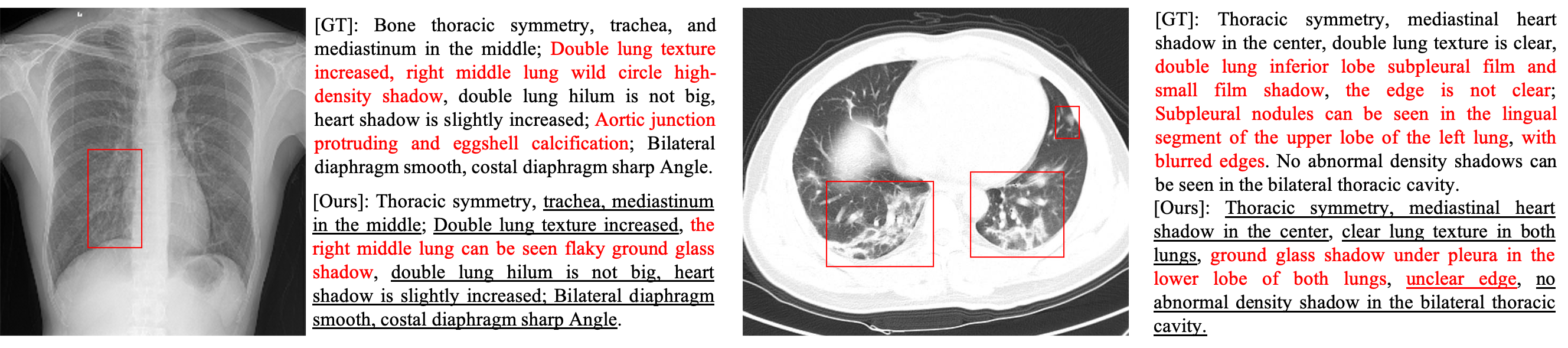}
\vspace{-.8 em}
\caption{Two samples from CX-CHR and COV-CTR datasets. Red bounding boxes annotated by a radiologist indicate the regions that he pays more attention to describing this image. The red text describes the abnormalities. Underlined text indicates alignment between ground truth reports and generated reports.}
\vspace{-1.8 em}
\label{motivation}
\end{figure}

Generally speaking, when a radiologist describes a medical image, he/she will carefully inspect the abnormal regions after quickly browsing the global image, then write a report that draws on the knowledge he/she learned from external medical domain information and his/her working experience. However, unlike radiologists' working patterns, most existing methods\cite{wang2017chestx,kumar2018boosted, wang2018tienet} employ the global image as input and train their language model with the datasets' corpora only, both of which are limitations. As shown in Figure~\ref{motivation}, the attention regions take up only a small portion of the global image, but have been treated equally to other regions in previous works. Therefore, other regions could be considered as irrelevant noise that distract the model. Furthermore, as the attention regions differ between each image, it is difficult to manually crop them in pre-processing. Moreover, unlike common and trivial normality in medical reports, abnormalities are rare and diverse. Training with datasets' corpora thus makes it difficult to alleviate data deviation.

Accordingly, to mimic the behavior of medical experts and address the above-mentioned learning difficulties, we introduce two kinds of auxiliary signals: namely the internal fusion features, and external medical linguistic information. More specifically, we attend the auxiliary region features to global visual features in order to produce the internal auxiliary signal, while the external auxiliary signals are contained in a large-scale easily-accessed medical textbook. We are inspired by the recent great progress made in large-scale unsupervised/self-supervised vision and language understanding~\cite{sun2019unsupervised,zhu2019vision,xiong2019pretrained,devlin2018bert,radford2018improving} which has demonstrated that auxiliary signals can improve data efficiency during training and reduce the gap between the visual and linguistic domains. Due to the difficulty associated with acquiring and annotating medical images, these two auxiliary signals are comparatively far easier to access and can avoid data inefficiency.

Capitalizing on these new auxiliary signals, we propose an Auxiliary Signal-Guided Knowledge (ASGK) approach to guide knowledge encoding and natural language decoding in order to facilitate medical report generation. The medical graph decoder and natural language decoder are pretrained using external auxiliary signals, enabling them to memorize and phrase medical knowledge, while the internal signals facilitate the graph encoding that permits the incorporation of prior medical knowledge and bridging of the visual and linguistic information. To tackle the imbalance between normal and abnormal tag distributions, moreover, we adopt focal loss~\cite{lin2017focal} as our training strategy for tag classification. 

We further introduce a new COVID-19 CT Report (COV-CTR) dataset for use in validating the robustness and generalization ability of ASGK. Since December 2019, the novel COVID-19 virus has caused a global pandemic and infected millions of people across 200 countries. A key step in controlling the infection is that of identifying infected people. In addition to the Reverse Transcription Polymerase Chain Reaction (RT-PCR) tests, lung CT scan analysis has emerged as another essential testing method. Therefore, an accurately written report could assist patients and doctors to understand their health condition. We invited three radiologists with more than five years of working experience to apply their diagnostic skills to the public COVID-CT dataset\cite{zhao2020COVID-CT-Dataset} and use this information to construct the COV-CTR dataset.

We test our approach on the large-scale CX-CHR dataset and our COV-CTR dataset. We adopt CIDER-D~\cite{vedantam2015cider}, ROUGE-L~\cite{lin-2004-rouge} and BELU~\cite{papineni-etal-2002-bleu} as the metrics for evaluating our approach. Comprehensive experiments demonstrate that ASGK improves performance in terms of both tag classification and report generation. Our ablation studies also provide insight that enables us to determine how ASGK works well.

The main contributions of this paper are three-fold as follows:
\vspace{-.5 em}
\begin{itemize}
\item We identify and produce two kinds of auxiliary signals, namely the internal fusion visual features and the external medical linguistic information to facilitate graph encoding and medical knowledge learning respectively.
\item We design a medical tag graph encoder to transfer input features into higher-level information and adopt Generative Pre-Training (GPT)~\cite{radford2018improving} as our natural language decoder to generate accurate and robust medical reports.
\item We invite three radiologists with more than five years of experience to apply their diagnostic skills to the COVID-19 CT images~\cite{zhao2020COVID-CT-Dataset} and use this information to construct a new medical report dataset, COVID-19 CT Report which will be available.
\end{itemize}

\section{Related Work}

\textbf{Visual Captioning and Medical Report Generation.} Due to the rapid development of deep learning~\cite{huang2017densely,he2016deep,hochreiter1997long,chung2014empirical}, visual captioning models~\cite{anderson2018bottom,gan2017semantic,vinyals2015show} have achieved significant progress in summarizing visual information (images or videos) in a single sentence or topic-related paragraph. Neural encoder-decoder frameworks~\cite{sutskever2014sequence} and attention mechanisms~\cite{ranzato2015sequence,you2016image} have achieved great performance in both natural images and the medical domain. To further boost accuracy, scene~\cite{johnson2015image} and knowledge graphs~\cite{li2019knowledge} are explored to replace the encoded vectors that are able to take advantage of detected nodes and their relationships. In order to alleviate textual data bias, reinforcement learning directly uses metric as a reward and optimizes these non-differentiable metrics~\cite{liu2017improved,rennie2017self}; however, it is poor at implicitly balancing the visual data bias.

\textbf{Medical Image Analysis with Auxiliary Signals.} Recent works~\cite{islam2017abnormality,shin2016learning} discussed the application of deep learning technologies to the field of medical image analysis. However, due to the difficulty associated with accessing and annotating medical images, many researchers have attempted to use self-supervised learning to loosen the requirements of training data. The core of self-supervised learning involves the design of various proxy tasks that provide auxiliary signals for training deep neural networks\cite{jing2020self}. Furthermore, auxiliary signals are widely applied as the basic structure for image analysis. Adopting auxiliary signals to guide training has advantages in terms of boosting model performance and improving model robustness. Zhuang \textit{et al.}~\cite{zhuang2019self} found that auxiliary signals are likely to benefit 3D neural networks for brain hemorrhage classification and brain tumor segmentation. 

\textbf{Language Model Pre-training.} Natural language decoders are another critical part of the image captioning process. Recent breakthroughs in the field of pretrained language models, such as ELMO\cite{peters2018deep}, BERT\cite{devlin2018bert}, and XLNet\cite{yang2019xlnet}, have demonstrated the effectiveness of auxiliary signals for a widespread range of natural language processing tasks. For example, the new state-of-the-art GPT-2\cite{radford2019language} reveals that pretraining allows models to learn a language's syntactic and semantic information via unsupervised learning, which is then transferred to other tasks. However, directly applying these models to medical domain datasets often yields unsatisfactory results due to the existence of a domain gap between general corpora and medical corpora. To tackle this problem, Habibi \textit{et al.}~\cite{habibi2017deep} proposes a completely generic method based on deep learning and statistical word embedding, while Lee \textit{et al.}~\cite{lee2020biobert} pretrains BERT on medical corpora.

\section{Approach}
\vspace{-.5 em}
\subsection{Problem Setup}

The task of medical report generation involves asking a model to generate a topic related paragraph consisting of a series of sentences to describe a medical image of a patient case. We represent the image as $I$ and the report as $S=\left\{w_1, w_2,..., w_l|w_i\in \textbf{V}\right\}$, where $w_i$ presents the index of word in $\textbf{V}$ the vocabulary of all words contained in the datasets. To generate fine-grained and semantically coherent sentences, we propose a graph encoder-decoder framework that first encodes inputs feature vectors to a medical tag graph and then decodes them to a medical report. We represent the medical tag as $G=(V, E)$, where $V=\left\{v_i\right\}_{i=1:N_t}$ and $E=\left\{e_{i,j}\right\}_{i,j=1:N_t}$ is a set of edges. In our task, we represent each node feature $v_i$ by its detected tag classification probability, then encode the correlation between each of the two tags as edge weights. $N_t$ represents the total number of medical tags composes abnormal terminologies, such as ``pneumothorax'' and ``colon shadow'', and normal terminologies such as ``normal spine'', ``normal intercostal space'' and so on.

Generally, when a radiologist describes a image, he will inspect the abnormal region carefully after quickly browsing the global image, then write a report that reflects both his inspection and the knowledge obtained from external medical domain information and his working experience. To mimic this pattern, we firstly pretrain the framework with the external medical signals collected from an appropriate website in order to correctly phrase and learn medical knowledge. Subsequently, the internal visual fusion signals facilitate graph encoding and bridge the gap between linguistic and visual domain. More details regarding these internal visual fusion signals are described in Section~\ref{asgl}.

\subsection{The structure of ASGK}

\begin{figure}[t]
\centering
\includegraphics[width=1\textwidth]{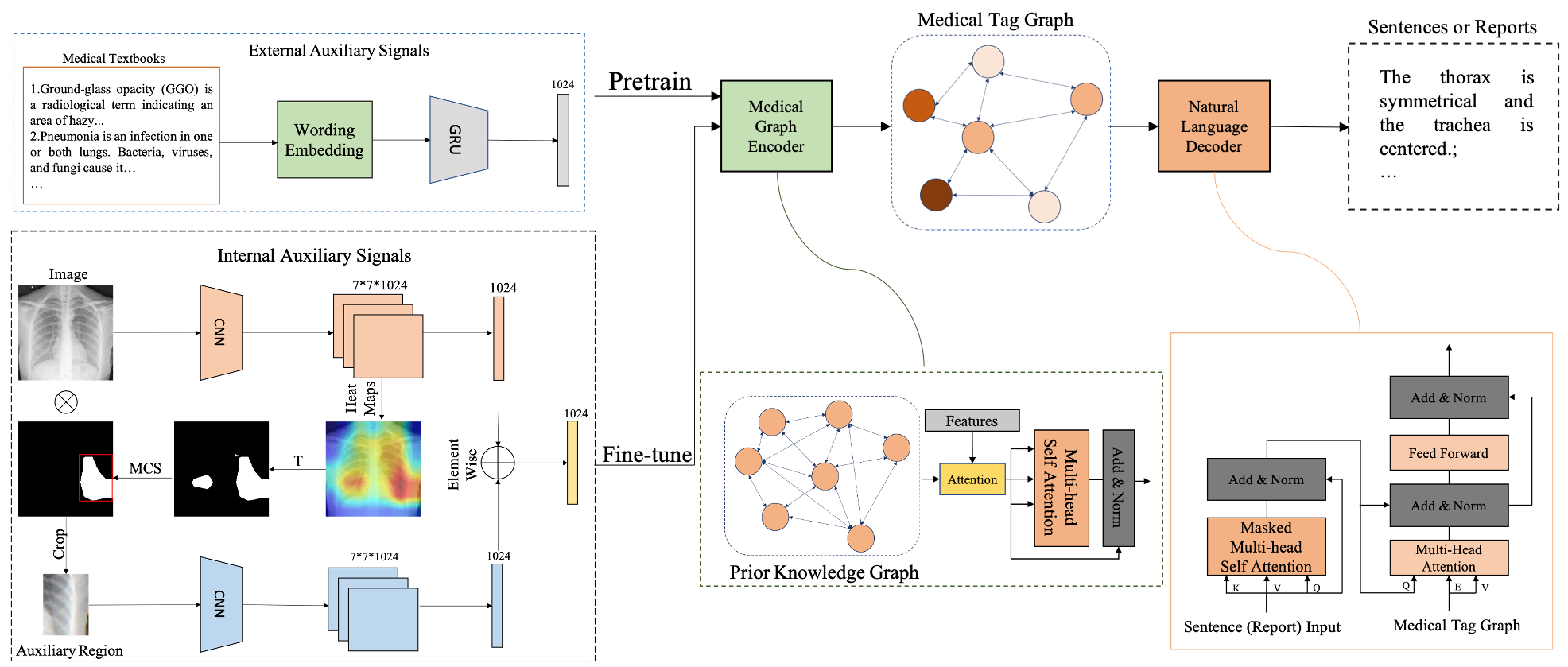}
\vspace{-1.0 em}
\caption{An overview of our ASGK approach. The ASGK model consists of a medical graph encoder and a natural language decoder. The medical graph encoder encodes input features into the corresponding medical tag graph, while the natural language decoder transfers high-level information to sentences or reports. The external signals guide the pretraining procedure, while the internal signals guide the model to bridge linguistic and visual information. T and MCS represent threshold and max connection select operation respectively.}
\vspace{-1.0 em}
\label{overview}
\end{figure}

An overview of our approach is shown in Figure~\ref{overview}. The main structure of ASGK comprises a medical graph encoder and natural language decoder.

\textbf{Medical Graph Encoder.} This component is built to encode the input features into higher level information, \textit{i.e.} a medical tag graph. In the medical graph, each node denotes one detected medical tag, the features of which are the classification probabilities and can be written as Equation~\ref{v}. 
\begin{align}
    V = \rm{Sigmoid}(W_{v}f_{input}) \label{v}
\end{align}
where $W_{v}$ is a projection matrix of size $N\times{d}$; here, $d$ represent the dimension of the input features, and N is the number of total tags. Given that the truth edge information is not available in our case, we conduct an attention operation to learn edge weights automatically, which can be written as follows:
\begin{align}
    e_{i,j} = \rm{Norm}(\rm{Attention}(W_{v}v_i, W_{v}v_j))
\end{align}
where \textit{Norm} is the normalization operation, while \textit{Attention} is executed as a scaled dot-product operation. Then the medical tag graph is incorporated with the prior medical knowledge which is represented as a set of nodes of size N with initialized features and edges via attention mechanism following by \cite{li2019knowledge}, which can be written as follows:
\begin{align}
    G = \rm{att}(G_{prior}, V, E)
\end{align}
To enhance the correlation between each of the nodes, we employ a multi-head self attention operation on $G$ to get the final graph. We further treat medical tag detection as a multi-label classification task and adopt BCE loss to maximize the prediction scores 
\begin{align}
    L_{tagcls} = -\sum^{N-1}_{i=0}y_i\log{v_i}+(1-y_i)\log(1-v_i) \label{tc}
\end{align}
where $W_{v}$ is a projection matrix of size $N\times{d}$; here, $d$ represent the dimension of the input features, $y_i$ is the ground truth label, and $v_i$ is the final graph tag features.

\textbf{Natural Language Decoder.}
Inspired by GPT~\cite{radford2018improving}, we design a natural language decoder consisting of $N=3$ blocks, similar to the Transformer decoder, to interpret the medical tag graph and enable semantic alignment in the visual and linguistic domain. The structure of the block is presented in Figure~\ref{overview}. This block applies a masked, multi-head self-attention operation to the medical report or sentences tokens $T=\left\{t_1, t_2,..., t_l\right\}$ embedded from Glove vectors pretrained on our datasets. We use \cite{radford2018improving} to maximize the likelihood in the following formulation:

\begin{align}
    L_t(T) = -\sum_{i}\log{P(t_i|t_1,...,t_{i-1}; \Theta)}
\end{align}
where $P$ is the conditional probability of the next token prediction, modeled using a neural network with parameters $\Theta$ and history sentences. Then, followed by position-wise feed forward layers, the natural language decoder aims to produce an output distribution over all token vocabulary.
\begin{align}
    h_0 =& I_WW_e+I_PW_p, \\
    H_l =& \textbf{block}(h_{l-1}, V, E) \forall{l}\in [1, N], \\
    P_i =& \rm{Softmax}(h_NW^T_e)
\end{align}
where $I_W$ is the index of input tokens in the vocabulary, $I_P$ is the index of the token's position, $W_e$ is the pretrained wording embedding matrix, and $W_p$ is the position embedding matrix. 

\subsection{Auxiliary Signal-Guide Learning}
\label{asgl}

\textbf{Pretraining with External Auxiliary Signals.} The direct application of general pretrained language models to medical domain tasks leads to unsatisfactory results, since the word distributions differ from those of those of general and medical corpora. To resolve this problem, we collect medical textual information from an appropriate website to construct a large-scale medical textbook. This textbook provides sufficient information about medical knowledge, including the symptoms, manifestations and other information about COVID-19 and thoracic diseases. Before feeding it into the medical graph encoder, we divide the medical textbook into sentences and embed the word tokens with embedding vectors, which are trained in our datasets using Glove. After embedding, sentences are encoded using a single-layer GRU with 1024 hidden units to produce the external medical auxiliary signals.

\textbf{Training with Internal Auxiliary Signals.} Evidently, the quality of the encoded medical graph will significantly affect the accuracy of the generated reports. Therefore, we produce internal fusion visual signals to facilitate medical graph encoding and bridge the gap between linguistic and visual information. As shown in Figure~\ref{overview}, we first classify the global image using DenseNet-121 and obtain the feature maps $ f_{c}\in{R^{7*7*1024}} $ before the final pooling layers and output from last pooling layers $f_{g}\in{R^{1*1024}}$. To produce the mask, we perform a threshold operation on a heat map acquired by Equation~\ref{heatmap} and select the max connected area:

\begin{equation}
\label{heatmap}
H = \mathop{\rm{max}}\limits_{k}(|f_{c}^{k}|), k\in{1:2048}
\end{equation}
We adopt another DenseNet to extract the attended region features $f_{l}\in{R^{1*1024}}$ from the final pooling layers, then perform the element-wise operation on $f_g$ and $f_l$ to produce the fusion signals $f_f$. To balance the deviation in medical tags, we optimize the parameters of three branch via focal loss, as follows:

\vspace{-2. em}
\begin{align}
    p^*_i = 
\left\{
\begin{array}{cc}
    p_i,&\;{\rm if\;\;\;y_i=1}  \\
    1-p_i,&\;{\rm otherwise} 
\end{array}
\right.
\end{align}

\vspace{-2. em}
\begin{align}
    L_{focal} = -\sum^{N-1}_{i=0}\alpha(1-p^*_i)^\gamma\log{p^*_i}
\end{align}
where $y_i$ represents the label, $p_i$ represents the prediction probability, $\alpha$ is a hyper-parameter set according to diverse datasets, and $(1-p^*_i)^\gamma$ is treated as a modulating factor with a tunable focusing parameter $\gamma\geq0$. We set $\alpha$ to $0.25$ and $\gamma$ to $2$ in our task.

\section{Experiments}

\begin{figure}[t]
\centering
\includegraphics[width=1\textwidth]{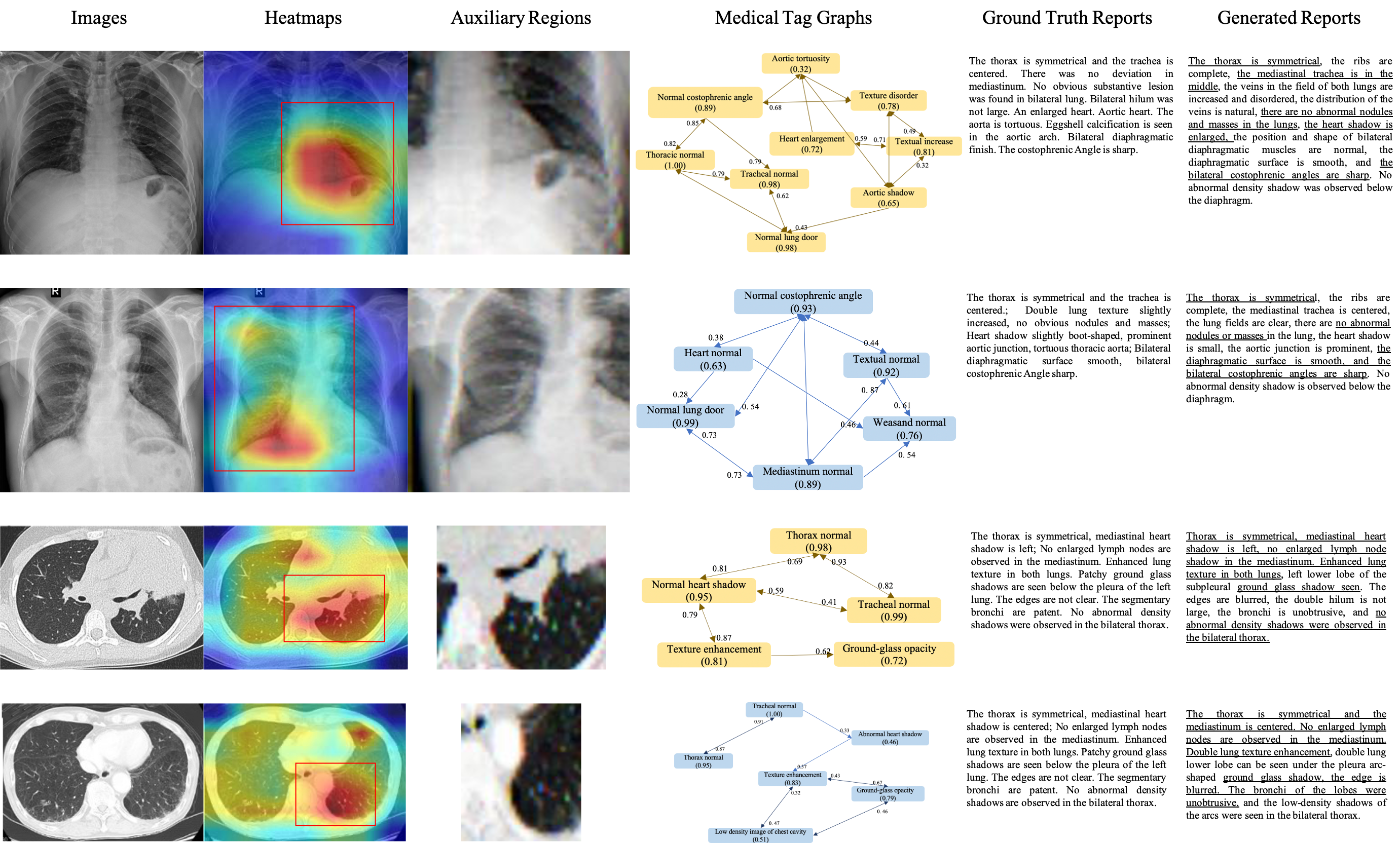}
\vspace{-1.0 em}
\caption{Sample output of our approach on both CX-CHR and COV-CTR datasets. We use the outputs before the last pooling layer in DenseNet-121 to generate heat maps, then threshold them by $\tau=0.7$ to produce the auxiliary regions. In the medical tag graphs, we show the nodes whose value (which is equal to the classification probability) exceeds $0.5$ and edges whose weights are more than 0.3. To read the image clearly, we show the values of some edges in the appropriate places. The underlined text indicates alignment between ground truth reports and generated reports.}
\vspace{-1.0 em}
\label{vresults}
\end{figure}

\textbf{Datasets.} We conduct experiments on the large-scale CX-CHR dataset and our COVID-19 CT Report dataset in order to validate the robustness and generalization ability of ASGK. CX-CHR is a large-scale chest X-ray dataset, constructed by a professional medical institution, that consists of 35,609 patients and 45,598 images paired with their corresponding Chinese diagnostic reports. We collect 173 medical tags comprising 155 abnormal terminologies and 28 normal terminologies from the 'findings' section and annotate paired images with these tags. Moreover, the COV-CTR datasets consist 728 images (349 for COVID-19 and 379 for Non-COVID) collected from published papers and their corresponding paired Chinese reports. We perform the same operation described above and collect 68 tags (50 abnormalities and 18 normalities).

We tokenize all reports and the medical textbook and filter tokens with a minimum frequency of three, which results in 27683 unique tokens covering over $98.7\%$ of words in the corpus. On both datasets, we randomly split the data into training, validation, and testing sets using a ratio of $7:1:2$; there is no overlap between these branches.

\textbf{Evaluation Metrics.} Following \cite{li2019knowledge}, we adopt three kinds of metrics to evaluate our approach. Firstly, we use area under the curve (AUC) to evaluate the performance of all medical tag classifications. Moreover, to evaluate medical report generation, we select CIDER-L, ROUGE-L, and BELU as automatic metrics and conduct the human evaluation. We randomly select 100 samples from the testing set and generate corresponding medical reports using CoAtt~\cite{jing2017automatic} and our approach.

\textbf{Training Details.} The whole network is implemented using a PyTorch framework based on Python 3.6 and trained on two GeForce RTX 2080T GPUs. We adopt DenseNet-121 with no pretraining as the backbone to extract visual features. There are three steps in our training process: external auxiliary signal-guide pretraining, DenseNet pretraining, and internal auxiliary guide training. In the first step, the maximum length of the sentence is 300 (padded with 0s), and the word embedding dimension is 300. We train ASGK for 30 epochs until convergence. The natural language decoder consists of three blocks. We adopt ADAM for optimizing and the training rate is 5e-4. For the second step, we resize the image to $224\times224$ for both global and region images. The batch size is 32. We jointly train two DenseNets for 50 epochs until convergence. The learning rate starts from 1e-2 and delays by $0.1$ every 10 epochs until 1e-5. We threshold the heat map by 0.7 to acquire region images. We adopt the model that achieves the best performance on test datasets as a visual extractor in the third step. In the final step, we resize the images to $224\times224$ and train the entire network for 30 epochs until convergence. The learning rates for the visual extractor and ASGK are 1e-5 and 5e-4, respectively. We also adopt the ADAM optimizer to minimize the loss function. Among the multi-tasks, we set all loss weights to 1.

\vspace{-.8 em}
\subsection{Results and Analysis}
\vspace{-.8 em}

\begin{table}[t]
\centering
\caption{Evaluation metrics on CH-CHR and COV-CTR datasets comparing ASGK with other methods. C and R are shot for CIDER-D and ROUGE-L. B-n denotes that the BLEU score uses up to n-grams. Hit represents the human evaluation results.}
\label{results}
\begin{tabular}{c|c|ccccccc}
\hline
Dataset                  & Model                                             & C                               & R                              & B-1                            & \multicolumn{1}{l}{B-2}        & \multicolumn{1}{l}{B-3}        & \multicolumn{1}{l}{B-4}        & \multicolumn{1}{l}{Hit(\%)}    \\ \hline
\multirow{6}{*}{CX-CHR}  & CoAtt\cite{jing2017automatic}    & 273.5                           & \textbf{64.5} & 64.7                           & 57.5                           & 52.5                           & 48.7                           & 8.0                            \\
                         & HRGR-Agent\cite{li2018hybrid}    & 289.5                           & 61.2                           & 67.3                           & 58.7                           & 53.0                           & 48.6                           & -                              \\
                         & KERP\cite{li2019knowledge}       & 285.0                           & 61.8                           & 67.3                           & 58.8                           & 53.2                           & 47.3                           & -                              \\
                         & Vision-BERT\cite{devlin2018bert} & 302.4                           & 63.7                           & \textbf{68.6} & 60.1                           & 54.1                           & 50.3                           & 19.0                           \\
                         & Vison-GPT\cite{radford2018improving}                                         & 301.8                           & 63.0                           & 67.9                           & 59.6                           & 54.0                           & 48.7                           & -                              \\
                         & Ours                                              & \textbf{324.5} & 64.1                           & \textbf{68.6} & \textbf{60.8} & \textbf{55.8} & \textbf{52.3} & \textbf{20.0} \\ \hline
\multirow{6}{*}{COV-CTR} & CoAtt\cite{jing2017automatic}    & 67.2                            & \textbf{74.8} & 70.9                           & 64.5                           & 60.3                           & 55.2                           & 25.0                           \\
                         & SAT\cite{vinyals2015show}        & 65.9                            & 72.3                           & 69.7                           & 62.1                           & 56.8                           & 51.5                           & -                              \\
                         & AdaAtt\cite{lu2017knowing}       & 68.2                            & 72.6                           & 67.6                           & 63.3                           & 59.6                           & 51.4                           & -                              \\
                         & Vision-BERT\cite{devlin2018bert} & \textbf{68.4}  & 74.7                           & 71.0                           & 65.3                           & 60.6                           & 55.8                           & 26.0                           \\
                         & Vision-GPT\cite{radford2018improving}                                        & 68.0                            & 74.6                           & 70.8                           & 64.5                           & 60.0                           & 54.9                           & -                              \\
                         & Ours                                              & \textbf{68.4}  & 74.6                           & \textbf{71.2} & \textbf{65.9} & \textbf{61.1} & \textbf{57.0} & \textbf{27.0} \\ \hline
\end{tabular}
\vspace{-1.2 em}
\end{table}

\textbf{Automatic Evaluation.} Table~\ref{results} summarizes the performances on the automatic evaluation metrics of different models. The results on both datasets indicate that ASGK outperforms all existing state-of-the-art models through its exploitation of auxiliary signals to guide the framework in knowledge pretraining and knowledge transfer procedures. The results demonstrate the robustness and superior generalization ability of ASGK. We also combine our medical graph encoder with Vision Bert~\cite{devlin2018bert} and Vision GPT\cite{radford2018improving} in order to validate the capability of the language-to-vision transfer. We adopt CIDER-D as the main metric to validate our model. On the large-scale CX-CHR dataset, ASGK significantly boosts performance compared with other baselines, it increases the CIDER score by 51.0, 35.0, 39.5, 22.1 and 22.7 respectively. However, ASGK only acheives a slightly low ROUGE-L score than the CoAtt\cite{jing2017automatic} method. ASGK also outperforms other baselines in COV-CTR dataset.


\textbf{Medical Tags Classification.} The AUCs of medical tag classification, which contains both normal and abnormal terminologies on both datasets, are presented in Table~\ref{ablation study}. Our framework, which is guided by two auxiliary signals, outperforms the baseline on both datasets. Baseline outputs are predicted by a DenseNet-121 without pretraining. We attempt to boost the performance through the use of internal auxiliary signals and the adaptation of focal loss to balance the deviation. This demonstrates that internal auxiliary signals effectively promote the medical graph encoder and facilitate the medical tag classification.

\textbf{Human Evaluation.} Given 200 random images from these two datasets equally, we invited three radiologists to evaluate the corresponding outputs of our methods, CoAtt\cite{jing2017automatic} and Vison-Bert\cite{devlin2018bert}. They are encouraged to select a more accurate result from each pair. The human evaluation results are presented in Table~\ref{results}. It shows that in the CX-CHR and COV-CTR datasets, radiologists thought $20\%$, and $27\%$ portions of our reports are more accurate than others' respectively, and while they thought $53\%$, and $22\%$ portions of results are same. The human evaluation demonstrates that our method is capable of generating accurate and semantic-coherent reports.

\textbf{Visualization.} An illustration of heat maps, auxiliary regions, medical tag graphs, and paragraphs of medical reports is presented in Figure~\ref{vresults}. It is clear from the results that auxiliary regions suggest the region on which the model should focus. For example, in the first row, the auxiliary region focuses on the inferior lobe of the left lung which presents a shadow. In the fourth row, moreover, the auxiliary region focuses the inferior pleural of the left lung, which covers ground-glass opacity, one of the symptoms of COVID-19. The medical tag graph demonstrates that ASGK is capable of encoding input features into a high-level knowledge graph; as we lack the ground truth of the corresponding graph, we train in an end-to-end way to encode the graph. The generated reports demonstrate the high quality and provide significant alignment with the ground truth.

\subsection{Ablation Studies}

\begin{table}[t]
\centering
\caption{Ablation studies for different auxiliary signals. IA, EA and CE are short for ``internal auxiliary signals'', ``external auxiliary signals' and ``cross entropy''. Four metrics are adopted to evaluate our model on two datasets.}
\label{ablation study}
\begin{tabular}{c|c|cccc}
\hline
Dataset                  & Model          & CIDER-D & ROUGE-L & BLEU-4 & AUC  \\ \hline
\multirow{5}{*}{CX-CHR}  & baseline       & 289.7   & 61.3    & 48.3   & 78.7 \\
                         & baseline+IA+CE & 304.6   & 62.5    & 48.9   & 82.1 \\
                         & baseline+IA    & 305.3   & 62.7    & 49.1   & 83.2 \\
                         & baseline+EA    & 317.2   & 63.8    & 52.0   & 79.3 \\
                         & baseline+IA+EA & \textbf{324.5}   & \textbf{64.1}    & \textbf{52.3}   & \textbf{85.9} \\ \hline
\multirow{5}{*}{COV-CRT} & baseline       & 59.1    & 68.3    & 52.5   & 72.7 \\
                         & baseline+IA+CE & 61.3    & 70.2    & 54.1   & 79.0 \\
                         & baseline+IA    & 62.8    & 70.5    & 54.2   & 79.7 \\
                         & baseline+EA    & 66.9    & 72.0    & 55.6   & 74.5 \\
                         & baseline+IA+EA & \textbf{68.4}    & \textbf{74.6}    & \textbf{57.0}   & \textbf{80.4} \\ \hline
\end{tabular}
\vspace{-1. em}
\end{table}

We conduct ablation experiments to compare the performance of the two auxiliary signals. Table~\ref{ablation study} presents the results of automatic evaluation metrics and tag classification. The baseline represents the direct training of the ASGK model without any auxiliary signals. In addition to extra notes, we adopt focal loss as our training strategy.

\textbf{Do internal auxiliary signals help?} From Table~\ref{ablation study}, we can determine that auxiliary signals significantly boost the tag classification performance and improve the quality of generated reports. The internal auxiliary signal-guided learning outperforms the automatic metrics $15.6\%$, $1.4\%$ and $0.6\%$ respectively, and also performs $4.5\%$ better than the baseline in terms of classification accuracy on the CX-CHR dataset. The quality of the medical tag graphs significantly impacts the natural language decoder. We produce internal auxiliary signals to mimic radiologists' working patterns, since abnormal regions provide richer visual features. These experiments demonstrate that focusing on abnormal regions benefits the detection of medical tags and the generation of medical reports.

\textbf{What is the use of focal loss?} Radiologists are asked to describe all of their observations on one medical image, which leads to serious data deviation on medical tag labels and reports. Typically, each image contains three to five normal tags and a few abnormal terminologies. To alleviate the deviation in multi-label classification tasks, we adopt focal loss in order to optimize the parameters in DenseNet and the medical tag decoder. When the second and third rows are compared, the performance shows its capability to balance deviation and improve AUC metrics. Without focal loss, the performances on AUC metrics decrease by $0.9\%$ and $0.7\%$ respectively on the two datasets.

\textbf{Are external auxiliary signals useful?} The external auxiliary signals guide the pretraining procedure to assist the model in memorizing and phrasing medical knowledge. As expected, ASGK benefits a lot from the pretraining procedure. The performance on automatic metrics are boosted substantially from $289.7\%$ to $317.2\%$ and $59.1\%$ to $66.9\%$ on the two datasets respectively, which indicates that external auxiliary signal-guided training is capable of generating accurate and semantically coherent sentence. However, it improves the classification accuracy slightly, by $0.6\%$, and $1.8$ respectively on the two datasets, which demonstrates that exploiting medical domain knowledge primarily promotes the natural language decoder. Furthermore, our findings show that without external auxiliary signals, the model fails to alleviate the data bias and is therefore prone to repeating several specific words and sentences in one report. 

Overall, the internal signals mainly facilitate the medical tag encoder's effectiveness in generating fine-grained sentences and describing more medical tags. The external signals enable the natural language decoder to generate more semantically coherent sentences. 

\vspace{-.8 em}
\section{Conclusions and Future Work}
\vspace{-.8 em}

In this paper, we proposed an Auxiliary Signal-Guided Knowledge Encoder-Decoder approach that mimics radiologists' working patterns to generate fine-grained and semantically coherent medical reports. We investigated how to best crop the auxiliary region from the global medical image, how to exploit medical domain knowledge from medical textbook, and how these auxiliary signals work. Experiments demonstrate that ASGK outperforms existing methods and boosts the performance of medical report generation tasks on report generation and tag classification on two medical datasets. In the future, we plan to focus on building a general captioning framework guided by auxiliary signals to encode and decode general corpora knowledge.

\section*{Broader Impacts}

This work practically analyzes a meaningful task combined with the computer vision and natural language processing task, medical report generation. Especially when pandemic happens like COVID-19, robust and accurate medical report generation technology is of great clinical value, which can reduce the burden on doctors and enable people to more accurately grasp their health status. We propose an anthropomorphic model, mimicking radiologists' working patterns, to promote the medical report generation task via acquiring easily-accessed auxiliary signals. This approach may inspire those researchers who have limited access to medical image resources to dig deeper into adopting unsupervised learning methods to acquire more auxiliary signals to supervised this task and achieve state-of-the-art performances. However, it still needs more effort to provide theoretical interpretation for these auxiliary signals.

\section*{Acknowledgements}

Dr Chang is partially supported by Australian Research Council (ARC) Discovery Early Career Researcher Award (DECRA) under Grant no. DE190100626. The authors also would like to thank Prof Anyuan Li from The First Affiliated Hospital of Harbin Medical University for providing his domain knowledge.

\bibliographystyle{splncs04}
\bibliography{NIPS}

\begin{thebibliography}{10}
\providecommand{\url}[1]{\texttt{#1}}
\providecommand{\urlprefix}{URL }
\providecommand{\doi}[1]{https://doi.org/#1}

\bibitem{anderson2018bottom}
Anderson, P., He, X., Buehler, C., Teney, D., Johnson, M., Gould, S., Zhang,
  L.: Bottom-up and top-down attention for image captioning and visual question
  answering. In: {CVPR} (2018)

\bibitem{cao2018retrieve}
Cao, Z., Li, W., Li, S., Wei, F.: Retrieve, rerank and rewrite: Soft template
  based neural summarization. In: {ACL} (2018)

\bibitem{chung2014empirical}
Chung, J., Gulcehre, C., Cho, K., Bengio, Y.: Empirical evaluation of gated
  recurrent neural networks on sequence modeling. arXiv preprint
  arXiv:1412.3555  (2014)

\bibitem{devlin2018bert}
Devlin, J., Chang, M.W., Lee, K., Toutanova, K.: Bert: Pre-training of deep
  bidirectional transformers for language understanding. arXiv preprint
  arXiv:1810.04805  (2018)

\bibitem{gan2017semantic}
Gan, Z., Gan, C., He, X., Pu, Y., Tran, K., Gao, J., Carin, L., Deng, L.:
  Semantic compositional networks for visual captioning. In: Proceedings of the
  IEEE conference on computer vision and pattern recognition. pp. 5630--5639
  (2017)

\bibitem{habibi2017deep}
Habibi, M., Weber, L., Neves, M., Wiegandt, D.L., Leser, U.: Deep learning with
  word embeddings improves biomedical named entity recognition. Bioinformatics
  \textbf{33}(14),  i37--i48 (2017)

\bibitem{he2016deep}
He, K., Zhang, X., Ren, S., Sun, J.: Deep residual learning for image
  recognition. In: Proceedings of the IEEE conference on computer vision and
  pattern recognition. pp. 770--778 (2016)

\bibitem{hochreiter1997long}
Hochreiter, S., Schmidhuber, J.: Long short-term memory. Neural computation
  \textbf{9}(8),  1735--1780 (1997)

\bibitem{huang2017densely}
Huang, G., Liu, Z., Van Der~Maaten, L., Weinberger, K.Q.: Densely connected
  convolutional networks. In: Proceedings of the IEEE conference on computer
  vision and pattern recognition. pp. 4700--4708 (2017)

\bibitem{islam2017abnormality}
Islam, M.T., Aowal, M.A., Minhaz, A.T., Ashraf, K.: Abnormality detection and
  localization in chest x-rays using deep convolutional neural networks. arXiv
  preprint arXiv:1705.09850  (2017)

\bibitem{jing2017automatic}
Jing, B., Xie, P., Xing, E.: On the automatic generation of medical imaging
  reports. arXiv preprint arXiv:1711.08195  (2017)

\bibitem{jing2020self}
Jing, L., Tian, Y.: Self-supervised visual feature learning with deep neural
  networks: A survey. IEEE Transactions on Pattern Analysis and Machine
  Intelligence  (2020)

\bibitem{johnson2015image}
Johnson, J., Krishna, R., Stark, M., Li, L.J., Shamma, D., Bernstein, M.,
  Fei-Fei, L.: Image retrieval using scene graphs. In: Proceedings of the IEEE
  conference on computer vision and pattern recognition. pp. 3668--3678 (2015)

\bibitem{kumar2018boosted}
Kumar, P., Grewal, M., Srivastava, M.M.: Boosted cascaded convnets for
  multilabel classification of thoracic diseases in chest radiographs. In:
  {ICIAR} (2018)

\bibitem{lee2020biobert}
Lee, J., Yoon, W., Kim, S., Kim, D., Kim, S., So, C.H., Kang, J.: Biobert: a
  pre-trained biomedical language representation model for biomedical text
  mining. Bioinformatics  \textbf{36}(4),  1234--1240 (2020)

\bibitem{li2019knowledge}
Li, C.Y., Liang, X., Hu, Z., Xing, E.P.: Knowledge-driven encode, retrieve,
  paraphrase for medical image report generation. In: {AAAI} (2019)

\bibitem{li2018hybrid}
Li, Y., Liang, X., Hu, Z., Xing, E.P.: Hybrid retrieval-generation reinforced
  agent for medical image report generation. In: {NeurIPS} (2018)

\bibitem{lin-2004-rouge}
Lin, C.Y.: {ROUGE}: A package for automatic evaluation of summaries. In: Text
  Summarization Branches Out. pp. 74--81. Association for Computational
  Linguistics, Barcelona, Spain (Jul 2004),
  \url{https://www.aclweb.org/anthology/W04-1013}

\bibitem{lin2017focal}
Lin, T.Y., Goyal, P., Girshick, R., He, K., Doll{\'a}r, P.: Focal loss for
  dense object detection. In: {ICCV} (2017)

\bibitem{liu2017improved}
Liu, S., Zhu, Z., Ye, N., Guadarrama, S., Murphy, K.: Improved image captioning
  via policy gradient optimization of spider. In: Proceedings of the IEEE
  international conference on computer vision. pp. 873--881 (2017)

\bibitem{lu2017knowing}
Lu, J., Xiong, C., Parikh, D., Socher, R.: Knowing when to look: Adaptive
  attention via a visual sentinel for image captioning. In: Proceedings of the
  IEEE conference on computer vision and pattern recognition. pp. 375--383
  (2017)

\bibitem{papineni-etal-2002-bleu}
Papineni, K., Roukos, S., Ward, T., Zhu, W.J.: {B}leu: a method for automatic
  evaluation of machine translation. In: Proceedings of the 40th Annual Meeting
  of the Association for Computational Linguistics. pp. 311--318. Association
  for Computational Linguistics, Philadelphia, Pennsylvania, USA (Jul 2002).
  \doi{10.3115/1073083.1073135},
  \url{https://www.aclweb.org/anthology/P02-1040}

\bibitem{peters2018deep}
Peters, M.E., Neumann, M., Iyyer, M., Gardner, M., Clark, C., Lee, K.,
  Zettlemoyer, L.: Deep contextualized word representations. arXiv preprint
  arXiv:1802.05365  (2018)

\bibitem{radford2018improving}
Radford, A., Narasimhan, K., Salimans, T., Sutskever, I.: Improving language
  understanding by generative pre-training. URL https://s3-us-west-2.
  amazonaws. com/openai-assets/researchcovers/languageunsupervised/language
  understanding paper. pdf  (2018)

\bibitem{radford2019language}
Radford, A., Wu, J., Child, R., Luan, D., Amodei, D., Sutskever, I.: Language
  models are unsupervised multitask learners. OpenAI Blog  \textbf{1}(8), ~9
  (2019)

\bibitem{ranzato2015sequence}
Ranzato, M., Chopra, S., Auli, M., Zaremba, W.: Sequence level training with
  recurrent neural networks. arXiv preprint arXiv:1511.06732  (2015)

\bibitem{rennie2017self}
Rennie, S.J., Marcheret, E., Mroueh, Y., Ross, J., Goel, V.: Self-critical
  sequence training for image captioning. In: {CVPR} (2017)

\bibitem{shin2016learning}
Shin, H.C., Roberts, K., Lu, L., Demner-Fushman, D., Yao, J., Summers, R.M.:
  Learning to read chest x-rays: Recurrent neural cascade model for automated
  image annotation. In: Proceedings of the IEEE conference on computer vision
  and pattern recognition. pp. 2497--2506 (2016)

\bibitem{sun2019unsupervised}
Sun, Y., Tzeng, E., Darrell, T., Efros, A.A.: Unsupervised domain adaptation
  through self-supervision. arXiv preprint arXiv:1909.11825  (2019)

\bibitem{sutskever2014sequence}
Sutskever, I., Vinyals, O., Le, Q.V.: Sequence to sequence learning with neural
  networks. In: Advances in neural information processing systems. pp.
  3104--3112 (2014)

\bibitem{vedantam2015cider}
Vedantam, R., Lawrence~Zitnick, C., Parikh, D.: Cider: Consensus-based image
  description evaluation. In: {CVPR} (2015)

\bibitem{vinyals2015show}
Vinyals, O., Toshev, A., Bengio, S., Erhan, D.: Show and tell: A neural image
  caption generator. In: Proceedings of the IEEE conference on computer vision
  and pattern recognition. pp. 3156--3164 (2015)

\bibitem{wang2017chestx}
Wang, X., Peng, Y., Lu, L., Lu, Z., Bagheri, M., Summers, R.M.: Chestx-ray8:
  Hospital-scale chest x-ray database and benchmarks on weakly-supervised
  classification and localization of common thorax diseases. In: {CVPR} (2017)

\bibitem{wang2018tienet}
Wang, X., Peng, Y., Lu, L., Lu, Z., Summers, R.M.: Tienet: Text-image embedding
  network for common thorax disease classification and reporting in chest
  x-rays. In: {CVPR} (2018)

\bibitem{wu2016encode}
Wu, Z., Cohen, R.: Encode, review, and decode: Reviewer module for caption
  generation. arXiv preprint arXiv:1605.07912  (2016)

\bibitem{xiong2019pretrained}
Xiong, W., Du, J., Wang, W.Y., Stoyanov, V.: Pretrained encyclopedia: Weakly
  supervised knowledge-pretrained language model. arXiv preprint
  arXiv:1912.09637  (2019)

\bibitem{xu2015show}
Xu, K., Ba, J., Kiros, R., Cho, K., Courville, A., Salakhudinov, R., Zemel, R.,
  Bengio, Y.: Show, attend and tell: Neural image caption generation with
  visual attention. In: {ICML} (2015)

\bibitem{yang2019xlnet}
Yang, Z., Dai, Z., Yang, Y., Carbonell, J., Salakhutdinov, R.R., Le, Q.V.:
  Xlnet: Generalized autoregressive pretraining for language understanding. In:
  Advances in neural information processing systems. pp. 5754--5764 (2019)

\bibitem{you2016image}
You, Q., Jin, H., Wang, Z., Fang, C., Luo, J.: Image captioning with semantic
  attention. In: Proceedings of the IEEE conference on computer vision and
  pattern recognition. pp. 4651--4659 (2016)

\bibitem{zhao2020COVID-CT-Dataset}
Zhao, J., Zhang, Y., He, X., Xie, P.: Covid-ct-dataset: a ct scan dataset about
  covid-19. arXiv preprint arXiv:2003.13865  (2020)

\bibitem{zhu2019vision}
Zhu, F., Zhu, Y., Chang, X., Liang, X.: Vision-language navigation with
  self-supervised auxiliary reasoning tasks. In: {CVPR} (2020)

\bibitem{zhuang2019self}
Zhuang, X., Li, Y., Hu, Y., Ma, K., Yang, Y., Zheng, Y.: Self-supervised
  feature learning for 3d medical images by playing a rubik’s cube. In:
  International Conference on Medical Image Computing and Computer-Assisted
  Intervention. pp. 420--428. Springer (2019)

\end{thebibliography}
\end{document}